\documentclass[letterpaper, 10 pt, conference]{ieeeconf}  
\pdfoutput=1

\IEEEoverridecommandlockouts                              

\usepackage{graphicx}
\usepackage{amsmath}
\usepackage{amsfonts}
\usepackage{bm}
\usepackage{booktabs}

\usepackage{color}
\usepackage{url}
\usepackage[]{hyperref}
\usepackage[dvipsnames]{xcolor}

\newcommand{\figref}{Figure~\ref}
\newcommand{\tabref}{Table~\ref}

\title{\LARGE \bf
Tracker: Model-based Reinforcement Learning for Tracking Control of Human Finger Attached with Thin McKibben Muscles
}

\author{Daichi Saito$^{1} {}^\ast$, Eri Nagatomo$^{1} {}^\ast$, Jefferson Pardomuan$^{1}$, and Hideki Koike$^{1}$  
\thanks{$^{1}$Authors are with Department of Computer Science,
        Tokyo Institute of Technology, Tokyo, Japan,
        {\tt\small saito.d.ah@m.titech.ac.jp}}%
\thanks{${}^\ast$These authors contributed equally.}%
}

\begin{document}


\maketitle
\thispagestyle{empty}
\pagestyle{empty}

\begin{abstract}
To adopt the soft hand exoskeleton to support activities of daily livings, it is necessary to control finger joints precisely with the exoskeleton. The problem of controlling joints to follow a given trajectory is called the tracking control problem. In this study, we focus on the tracking control problem of a human finger attached with thin McKibben muscles. To achieve precise control with thin McKibben muscles, there are two problems: one is the complex characteristics of the muscles, for example, non-linearity, hysteresis, uncertainties in the real world, and the other is the difficulty in accessing a precise model of the muscles and human fingers. To solve these problems, we adopted DreamerV2, which is a model-based reinforcement learning method, but the target trajectory cannot be generated by the learned model. Therefore, we propose Tracker, which is an extension of DreamerV2 for the tracking control problem. In the experiment, we showed that Tracker can achieve an approximately $81$\% smaller error than PID for the control of a two-link manipulator that imitates a part of human index finger from the metacarpal bone to the proximal bone. Tracker achieved the control of the third joint of the human index finger with a small error by being trained for approximately $60$ minutes. 
In addition, it took approximately $15$ minutes, which is less than the time required for the first training, to achieve almost the same accuracy by fine-tuning the policy pre-trained by the user’s finger after taking off and attaching thin McKibben muscles again as the accuracy before taking off.
\end{abstract}

\section{Introduction}
An exoskeleton plays an important role in controlling the human body in support of activities of daily livings (ADL), such as walking~\cite{pinto2020performance}, grasping and dexterous manipulation~\cite{gull2020review, du2021review}. Focusing on the hand exoskeleton, various gloves have been proposed for rehabilitation, support of manual tasks for handicapped people and dexterous skill acquisition. Compared to electrical actuators such as motors for an exoskeleton, soft actuators such as pneumatic artificial muscles~\cite{suzumori2015new, kurumaya2017design} have been showing an increasing trend owing to their compliance and light weight. Several studies have proposed soft hand exoskeletons using thin McKibben muscles for rehabilitation~\cite{wang2020sensor}, hand assist~\cite{koizumi2020soft} and musical skill acquisition~\cite{takahashi2020soft}.

\begin{figure}[t]
  \centering
  \includegraphics[width=\columnwidth]{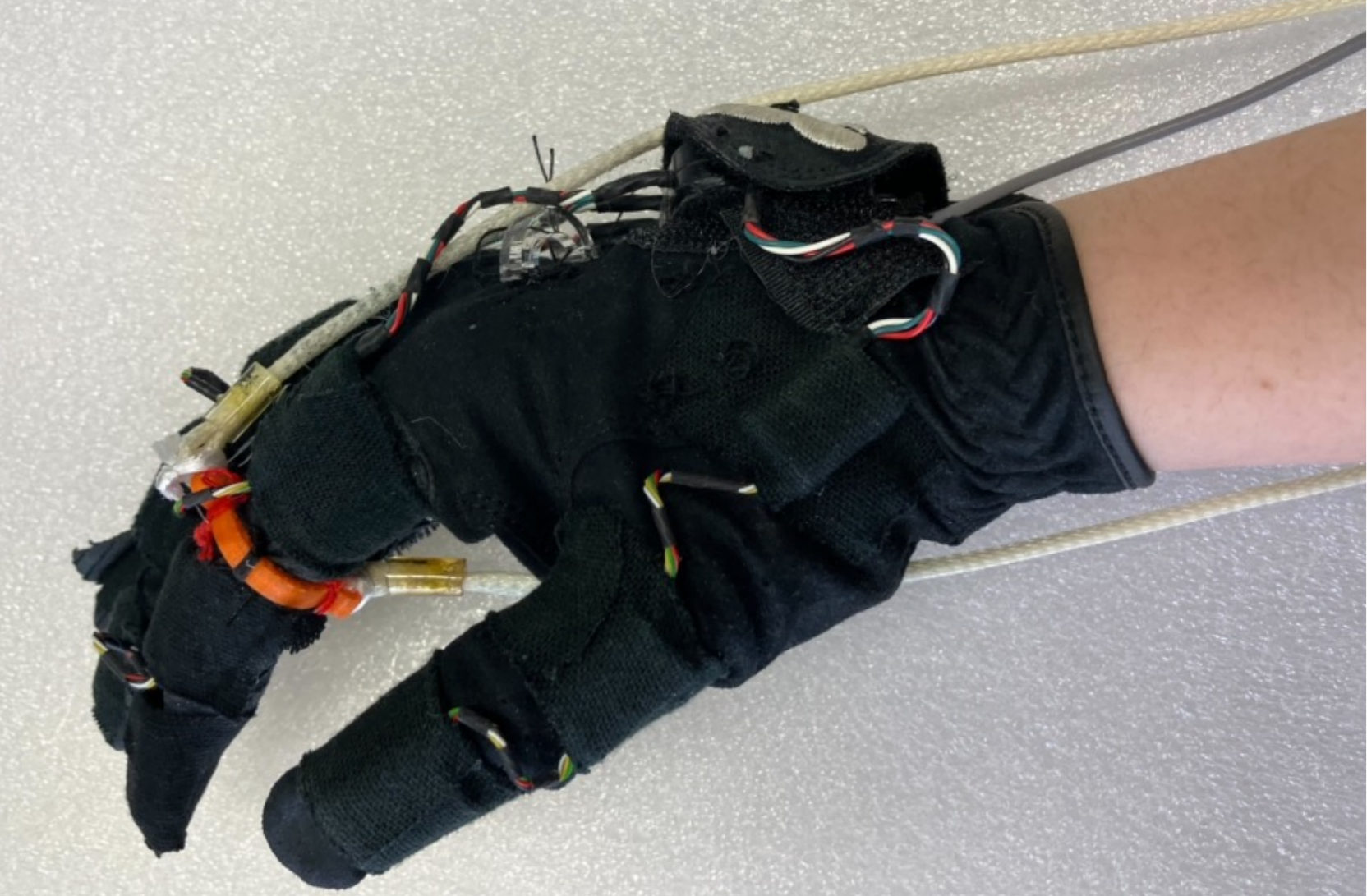}
  \vspace{-5mm}
  \caption{Human finger attached with thin McKibben muscles.}
  \label{fig:problem}
  \vspace{-3mm}
\end{figure}

To adopt soft hand exoskeletons to ADL support and skill acquisition, it is necessary to control finger joints precisely with soft hand exoskeletons. The problem of controlling joints to follow a given trajectory is called the tracking control problem.
It is difficult to precisely control with pneumatic artificial muscles because these soft actuators suffer from high non-linearity, hysteresis, which is a time-varying characteristic, and various uncertainties in the real world owing to rubber tube and fiber sleeve characteristics, complexity of air pressure, and friction. Because of the difficulty in control, previous studies have focused only on simple motions, such as opening and closing a hand and finger tapping. The tracking control problem for finger joints using a soft hand exoskeleton is very important for achieving dexterous finger motion, but designing a control policy is challenging and open-problem. In this study, we focus on the tracking control problem of human fingers attached with thin McKibben muscles (\figref{fig:problem}).

Many previous studies have used an open-loop control~\cite{wang2020sensor, koizumi2020soft} and feedback control, such as PID control~\cite{faudzi2016modeling, hafidz2022control, endo2020flexible}, as control methods for pneumatic actuators. Because it is difficult to formulate the dynamics of the fingers and pneumatic artificial muscles, an open-loop controller cannot achieve precise control. In addition, the PID controller requires manual parameter tuning. Finding the optimal control parameters is difficult because it requires a skillful technique. Recent advances in deep learning techniques have been adopted to control pneumatic artificial muscles. A convolutional neural network was adopted to control the deformation of the fabric actuator~\cite{yamaguchi2020three}. In \cite{yamaguchi2020three}, the convolutional neural network was trained using a randomly collected dataset. However, this study focused only on static deformation. Using this method in dynamic control requires an enormous amount of data because the cost function cannot be set explicitly. Reinforcement learning is a technique in which the cost function, which is called a reward, can be set and adopted for various control problems, such as games, robot grasping and manipulation~\cite{arulkumaran2017deep}. The muscle-skeleton robot control policy was trained using deep reinforcement learning~\cite{fan2020humanoid}. In \cite{fan2020humanoid}, model-free reinforcement learning was used. Model-free reinforcement learning requires enormous interaction with the environment; therefore, a physics simulator is required. Our focus is on the control of a human finger attached with thin McKibben muscles; therefore precise models of the human finger and thin McKibben muscles are needed. However, it is difficult to access these models. Thus, a data-efficient learning method is required to control human fingers attached with thin McKibben muscles.

\begin{figure}[t]
  \centering
  \includegraphics[width=\columnwidth]{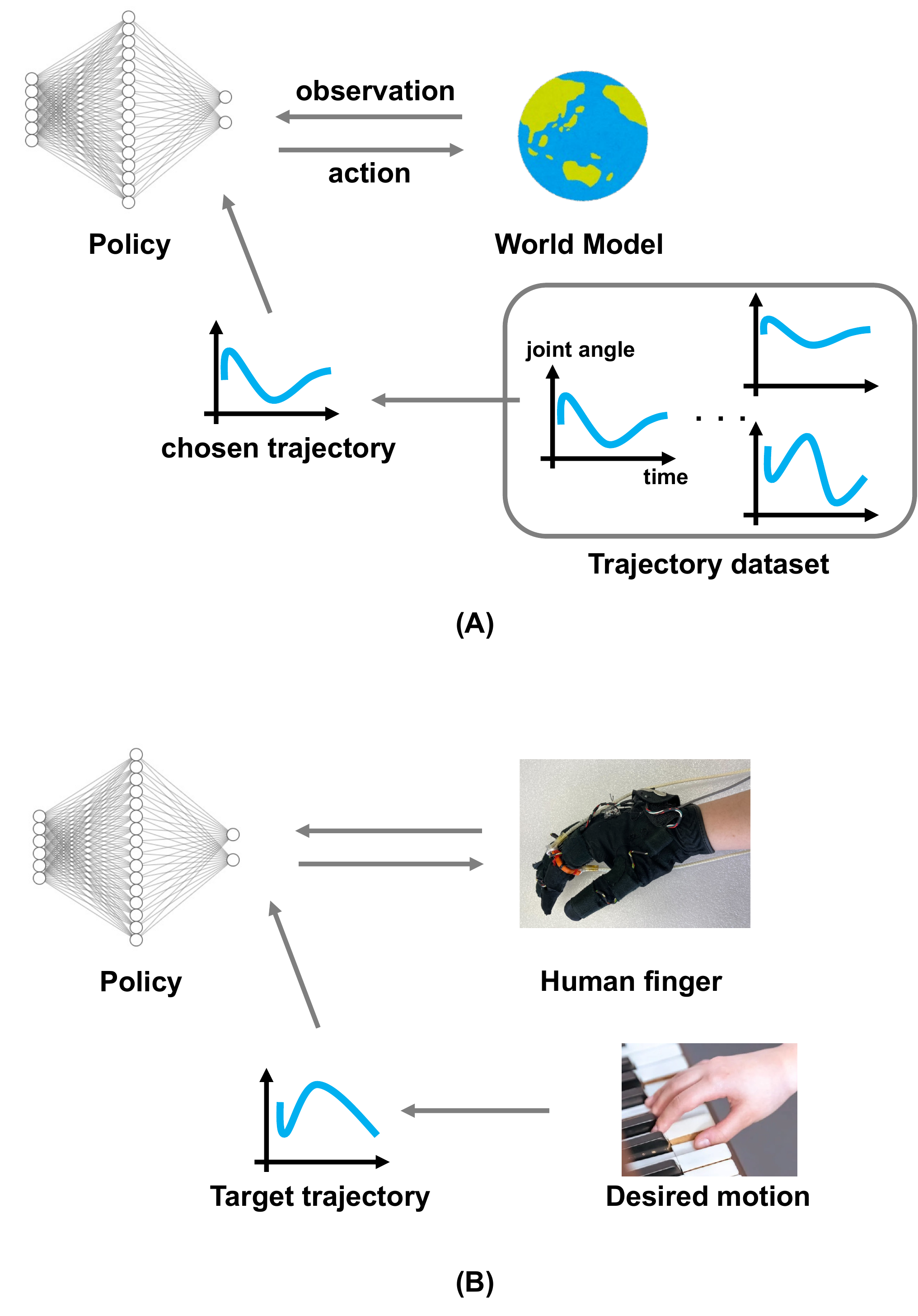}
  \vspace{-5mm}
  \caption{Overview of Tracker. (A) is the figure of the training phase, (B) is the figure of the execution phase. (A) Tracker trains the tracking control policy through the interaction with World Model. The policy decides the action from the observation generated by World Model and target trajectory. The target trajectory is randomly chosen from the pre-collected trajectory dataset. (B) During execution, the tracking control policy determines the action based on the observation of the human finger and the target trajectory. The target trajectory is the user's desired motion (for example, piano playing task in this figure).}
  \label{fig:concept}
  \vspace{-3mm}
\end{figure}

Model-based reinforcement learning is a data-efficient method for learning the control policies~\cite{arulkumaran2017deep}. Model-based reinforcement learning is a technique that learns the environment model from experience in the environment and train the policy from experience in the learned environment model. Recently, World Models~\cite{ha2018world} have been adopted to learn the precise model of the environment~\cite{hafner2020mastering}. DayDreamer~\cite{wu2022daydreamer} showed that it is possible to learn the policy of physical robot control using model-based reinforcement learning, DreamerV2~\cite{hafner2020mastering}. Inspired by \cite{wu2022daydreamer}, we show that model-based reinforcement learning is effective for learning the control policy of a human finger attached with thin McKibben muscles. In DayDreamer, the effectiveness of the model-based reinforcement learning on control soft actuators has not yet been investigated. In general, the control of soft actuators is more difficult than that of hard actuators owing to the complex characteristics of soft actuators. 
World Model cannot generate target trajectories to track because the target trajectory is given from the outside of the environment, not the environment. Note that the trajectory can be generated if it is included in the input of World Model, but the generated trajectory is not guaranteed to be the intended trajectory. This is why the tracking control policy cannot be learned just by adopting DreamerV2. Thus, we propose Tracker, a model-based reinforcement learning method for the tracking control problem (\figref{fig:concept}). Tracker is an extension of DreamerV2 to the tracking control problem. Tracker trains the tracking control policy through the interaction with World Model, which is learned from experience in the environment. The policy decides the action from the observation generated by World Model and the target trajectory randomly chosen from the pre-collected trajectory dataset. During execution, the tracking control policy determines the action based on the observation of the human finger and target trajectory. We can select any target trajectory within the range of the joint motion.

In this study, we focus on the tracking control problem of human fingers attached with thin McKibben muscles using model-based reinforcement learning. The contributions of this study are as follows:
\begin{itemize}
    \item We investigated the effectiveness of model-based reinforcement learning for controlling human fingers attached with thin McKibben muscles.
    \item We extended model-based reinforcement learning to adopt the tracking control problem.
    \item We demonstrated that it is possible to learn the tracking control policy with thin McKibben muscles using our extended model-based reinforcement learning in the experiment.
\end{itemize}

\section{Related Work}
An open-loop controller was used to control pneumatic artificial muscles in previous studies~\cite{wang2020sensor, koizumi2020soft}. An open-loop controller is easy to implement but is effective for simple motions (e.g., opening-closing fingers and finger tapping). It is difficult to adopt an open-loop controller for tracking control with pneumatic artificial muscles because it is difficult to formulate the dynamics of the actuators owing to their complex characteristics (e.g., high non-linearity, hysteresis). A feedback controller, such as a PID controller, is a model-free control method. Many previous studies have used PID controllers to control with pneumatic artificial muscles~\cite{faudzi2016modeling, hafidz2022control, endo2020flexible, buchler2016lightweight}. The PID controller is a general controller, but it requires the skilled technique to find the optimal control parameters. Control methods have been proposed to compensate for the slow response of pneumatic artificial muscles~\cite{suzuki2018novel, melkou2019high}. However, these methods require precise modeling of the actuators.

A deep neural network has been adopted to control pneumatic artificial muscles. A controller using a deep neural network does not require manual parameter tuning. A convolutional neural network was adopted to control the deformation of fabric actuators~\cite{yamaguchi2020three}. This study focused on the static deformation rather than dynamic deformation. In this study, this approach is called 'direct regression'. Direct regression cannot set the cost function explicitly and thus requires an enormous amount of data in dynamic control. Direct regression has been adopted for the dynamic control of flexible manipulators with a recurrent neural network~\cite{kawaharazuka2019dyamic}, however, \cite{kawaharazuka2019dyamic} focused on the simple dynamic motion.

Reinforcement learning is a technique that can set a cost function, which is called a reward. There are two types of reinforcement learning: model-free and model-based reinforcement learning. Model-free reinforcement learning requires enormous interaction with the environment, so a simulator is needed in general. In fact, \cite{fan2020humanoid} trained the policy of musculoskeletal robot control in a simulator. Additionally, the sim-to-real problem~\cite{zhao2020sim} should be considered. Model-based reinforcement learning requires a smaller amount of interaction than model-free reinforcement learning. In this study, we adopt model-based reinforcement learning because it is difficult to access a precise model of human fingers and pneumatic artificial muscles. There are various model-based reinforcement learning methods~\cite{ha2018world, Hafner2020Dream, hafner2020mastering}. It is possible to learn the policy of physical robot control using model-based reinforcement learning~\cite{wu2022daydreamer}. \cite{wu2022daydreamer} has shown that model-based reinforcement learning is capable of online learning in the real world. However, model-based reinforcement learning has not been shown to be effective in learning the control of soft robots. Generally, the control of soft robots is more difficult than that of hard robots. In this study, we investigate the effectiveness of model-based reinforcement learning on the control of thin McKibben muscles, one of the soft actuators.

\section{Method}
In this study, we focus on the tracking control problem of a human finger attached with thin McKibben muscles. The following two problems should be solved to achieve precise tracking control:
\begin{itemize}
    \item The complex characteristics of thin McKibben muscles, for example, non-linearity, hysteresis, which is a time-varying characteristic and uncertainties in the real world
    \item The difficulty in access to the precise model of thin McKibben muscles and human fingers
\end{itemize}
To solve the former problem, we use a deep-neural-network-based control policy that can deal with non-linearity, hysteresis and various uncertainties. To solve the latter problem, we require a data-efficient learning method because a real human finger and actuator must be used during the training of the control policy. Thus, we adopt model-based reinforcement learning, which is a data-efficient reinforcement learning. Inspired by the previous study~\cite{wu2022daydreamer}, we use DreamerV2~\cite{Hafner2020Dream}, which is World-Model-based reinforcement learning. However, the target trajectory is given from the outside of the environment, not the environment; therefore, the trajectory cannot be generated from World Model. Thus, we propose Tracker, which is an extension of DreamerV2 for the tracking control problem.

In this section, we describe DreamerV2 and Tracker, a model-based reinforcement learning method for the tracking control problem. We also describe the observation and trajectory, which is the target joint angles, action, reward.

\subsection{Preliminary: DreamerV2}
DreamerV2, which Tracker is based on, consists of Recurrent State-Space Model (RSSM)~\cite{hafner2019learning}, which is the component of World Model, and actor, critic. The RSSM is trained from a dataset of past experience. The experience dataset is grown through interactions with the actor in the environment. The actor and critic are trained from experience sequences generated by RSSM.

In detail, DreamerV2 consists of six components: RSSM (Reuccurent model, Representation model, Transition model), Observation predictor, Reward predictor, Actor, Critic. These components are conventionally represented by the following equations:
\begin{align}
\text{Recurrent model: } & h_t = f_{\phi}(h_{t-1}, z_{t-1}, a_{t-1})\\
\text{Representation model: } & z_t \sim q_{\phi}(z_t | h_t, x_t)\\
\text{Transition model: } & \hat{z}_t \sim p_{\phi}(\hat{z}_t | h_t)\\
\text{Observation predictor: } & \hat{x}_t \sim p_{\phi}(\hat{x}_t | h_t, z_t)\\
\text{Reward predictor: } & \hat{r}_t \sim p_{\phi}(\hat{r}_t | h_t, z_t)\\
\text{Discount predictor: } & \hat{\gamma}_t \sim p_{\phi}(\hat{\gamma}_t | h_t, z_t) \\
\text{Actor: } & \hat{a_t} \sim p_{\psi}(\hat{a}_t | \hat{z_t})\\
\text{Critic: } & v_{\xi}(\hat{z}_t) \approx E_{p_{\phi}, p_{\psi}}\left[\sum_{i\geq t}\hat{\gamma}^{i-t}\hat{r}_i\right]
\end{align}
$z_t$ and $h_t$ are stochastic and deterministic latent variables, respectively. $a_t$ is the action, $o_t$ is the observation from the environment, $\gamma_t$ is the probability of ending an episode. The actor and critic are learned from the stochastic latent variable $z_t$ and the reward $r_t$. 

The target trajectory is given from the outside of the environment, not the environment; therefore, the trajectory cannot be generated from RSSM. Note that the trajectory can be generated if it is included in the observation, but the generated trajectory is not guaranteed to be the intended trajectory. This becomes a problem when the range of joint motion is limited or when one joint motion is linked with the other joint motion. Thus, we propose Tracker, a model-based reinforcement learning method for the tracking control problem. 

\subsection{Tracker}
\begin{figure}[t]
  \centering
  \includegraphics[width=\columnwidth]{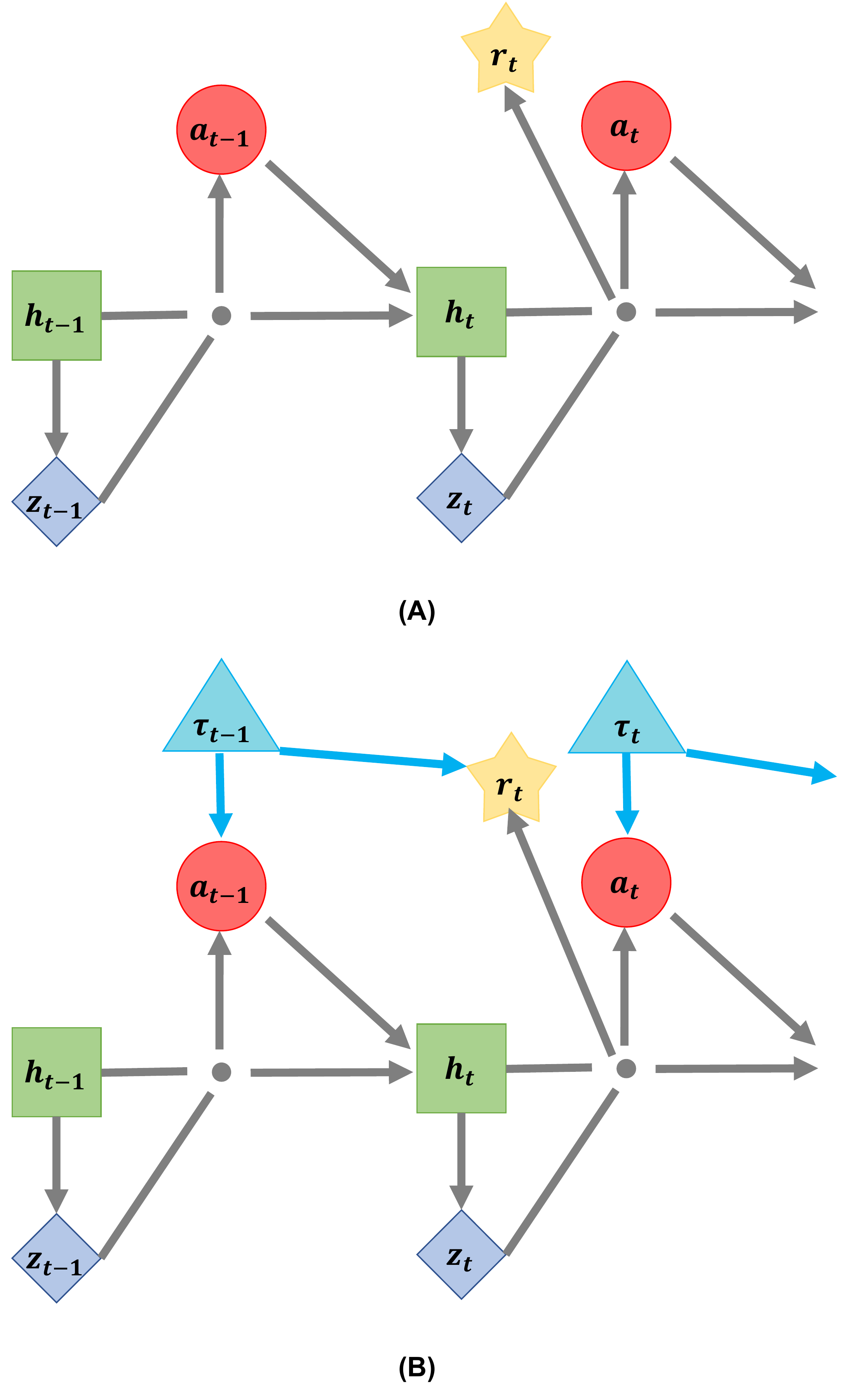}
  \vspace{-5mm}
  \caption{Comparison of the graphical model with (A) DreamerV2 and (B) Tracker. The transition of two latent variables $h_t, z_t$ are the same between DreamerV2 and Tracker. The action $a_{t-1}$ and reward $r_t$ conditioned by the target trajectory $\tau_{t-1}$ in Tracker.}
  \label{fig:extension}
  \vspace{-3mm}
\end{figure}

Tracker is an extension of DreamerV2 to the tracking control problem (\figref{fig:extension}). The target trajectory affects the reward and action, not the observation. Therefore, we condition Reward predictor and Actor, Critic using the target trajectory $\tau$. The Reward predictor, Actor, Critic are represented by the following equations:
\begin{align}
\text{Reward predictor: } & \hat{r}_t \sim p_{\phi}(\hat{r}_t | h_t, z_t,\tau_{t-1})\\
\text{Actor: } & \hat{a_t} \sim p_{\psi}(\hat{a}_t | \hat{z_t}, \tau_{t})\\
\text{Critic: } & v_{\xi}(\hat{z}_t, \tau_{t}) \approx E_{p_{\phi}, p_{\psi}}\left[\sum_{i\geq t}\hat{\gamma}^{i-t}\hat{r}_i\right]
\end{align}
The inputs of Actor and Critic are the same ($\hat{z}_t$ and $\tau_t$); therefore, it is possible to train the two models conventionally.

Actor and Critic are trained from the long-horizon pair data, which consists of the target trajectory and sequences generated by RSSM. During the learning of Actor and Critic, the target trajectory is randomly chosen from the prepared trajectory dataset, which meet the joint range of motion. Any trajectory that we wish to be learned by the control policy can be included in the trajectory dataset.

\subsection{Observation, Action, Trajectory, Reward}
\subsubsection{Observation}
Observation $o_t$ includes the angle and angular velocity for each joint, and the air pressure for each thin McKibben muscle. Let $n$ be the number of joints and $m$ be the number of actuators; then, $o_t \in \mathbb{R}^{2n+m}$. 

\subsubsection{Action}
Action $a_t$ includes the air pressure for each thin McKibben muscle; then, $a_t \in \mathbb{R}^m$.

\subsubsection{Trajectory}
Trajectory $\tau_t$ includes the angle and angular velocity for each joint up to $l$ steps ahead ($\tau_t \in \mathbb{R}^{2ln}$). By providing future trajectories, a more forward-looking action can be determined. The deep-neural-network based controller has the strength of including features other than the error compared with classical feedback controllers.

\subsubsection{Reward}
Reward $r_t$ is given by the following equation:
\begin{equation}
\begin{split}
    r_t = &-\sum_{i=1}^{n}\{(\frac{\|\theta_{i,t}-\hat{\theta}_{i,t}\|}{\alpha})^2+\frac{I_{i,t}}{\beta}\} -\sum_{i=1}^{m}(\frac{p_{i,t}}{\gamma})^2
\end{split}
\end{equation}
$\theta_{i,t}$ and $\hat{\theta}_{i,t}$ are the observable angles and target angles from $i$-th joint, respectively. $I_{i,t}$ equals $1$ if $\dot{\theta}_{i,t} \times \dot{\hat{\theta}}_{i,t} < 0$, otherwise $0$. $\dot{\theta}_{i,t}$ is the angular velocity of $i$-th joint and $\hat{\dot{\theta}}_{i,t}$ is the target angular velocity of $i$-th joint. $p_{i,t}$ denotes the air pressure in $i$-th actuator. The first term is used for the tracking control. To achieve smooth motion, $I_{i,t}$, which is a term for the angular velocity, is added. To avoid the burden on the joints and actuators due to the large air pressure, the second term is added.

\subsection{Implementation detail}
RSSM and Observation predictor, Reward predictor, Discount predictor, Actor, Critic are implemented as neural networks. Recurrent model uses a Fully Connected Layer and Gated Recurrent Unit (GRU)~\cite{cho2014learning}. The other components use just Fully Connected Layers. We use the exponential linear unit (ELU)~\cite{clevert2015fast} as the activation function for all components. We use the Adam optimizer~\cite{kingma2014adam}.

There are three processes in Tracker: World Model learning, Actor-Critic learning, Data collection from the real world. Running these processes synchronously causes slow learning because it takes a long time to collect data in a real world. This is a burden on the users. Thus, we implement asynchronous data collection for faster training with reference to \cite{zhang2020asynchronous}.

\section{Experiment I}\label{ex1}
To confirm whether our control policy, Tracker, can be applied to thin artificial muscles and to check the performance of our policy for the tracking control problem, we trained the policy on a two-link manipulator that imitates a human finger and compared Tracker with a PID controller.
\subsection{Settings}

\subsubsection{Hardware settings}
\begin{figure}[tb]
\begin{center}
    \includegraphics[width=1\columnwidth]{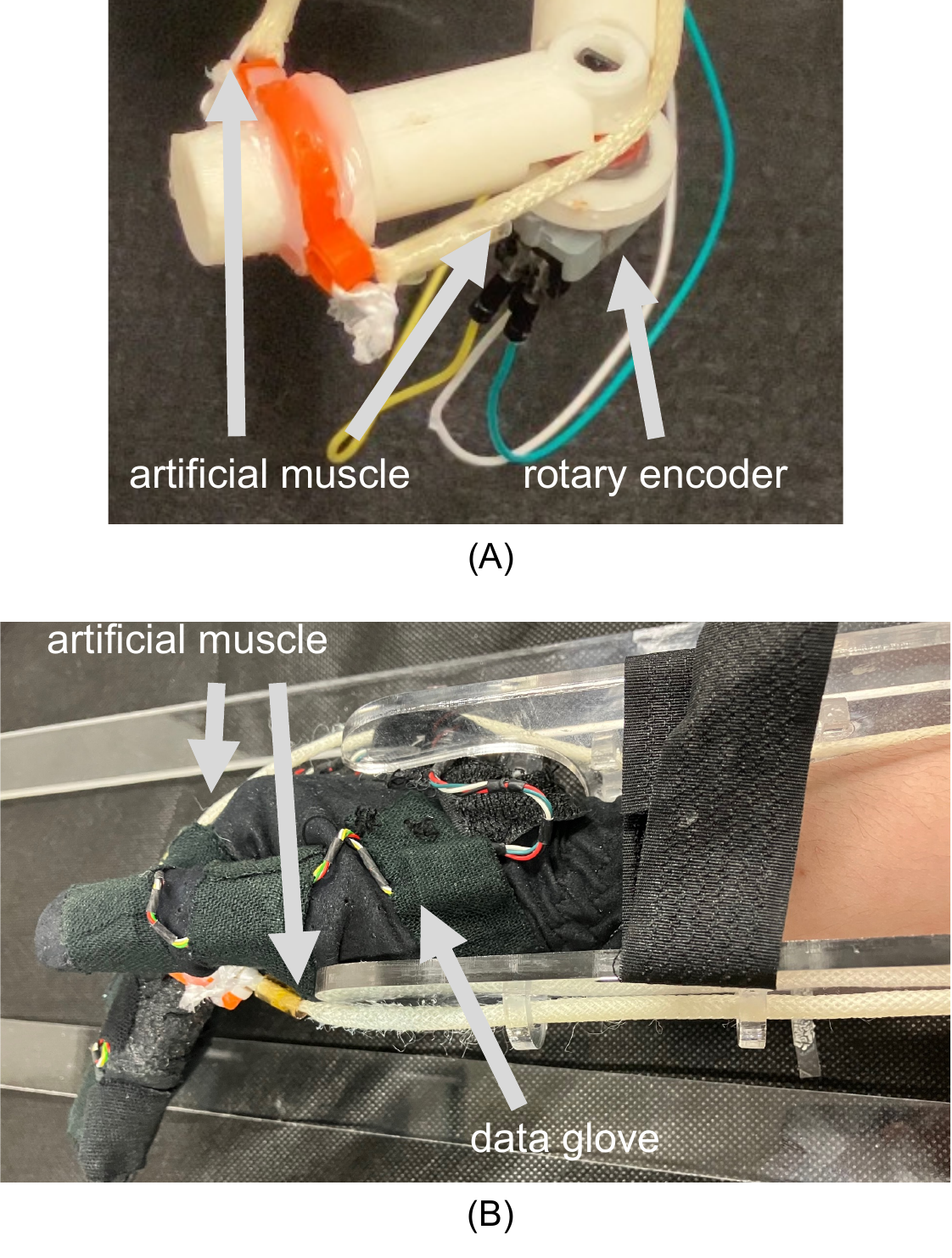}
    \caption{(A): Two-link manipulator that imitates a human finger. (B): Human finger attached with the artificial muscles. The hand is fixed using two acrylic plates.}
    \label{fig:platform}
\end{center}
\end{figure}

This experiment was performed on a two-link manipulator consisting of two links and one joint ($n=1$) that imitated part of the human index finger from the metacarpal bone to the proximal bone (\figref{fig:platform}-A). The two links were connected by a rotary encoder to obtain the joint angle. The joint angle was sent to the micro controller, ESP32-WROOM-32\footnote{https://www.espressif.com/en/products/modules/esp32}, for joint angle feedback. 

We used a thin McKibben artificial muscle with a diameter of $2.5$mm from s-muscle\footnote{https://www.s-muscle.com/}. A ring was attached to the tip of the first link, and two artificial muscles ($m=2$) were attached to this ring in an antagonistic state. 
The action, air pressure, was decided every $100$ms.

The parameters of the PID controller were determined using the Ziegler-Nichols method and then manually fine-tuned through some experimental trials.

\subsubsection{Training settings}

The length of experience sequence generated from RSSM was set to be $50$. Trajectory included the angle and angular velocity up to $3$ steps ahead ($l=3$).
Reward parameters were set as follows:
\begin{equation}
\begin{split}
    \alpha=40, \beta=1000, \gamma=50000
\end{split}
\end{equation}

The parameters of the reward were set as follows: Initially, the parameters were calculated to normalize the reward to the range from $-1$ to $1$ for each term. Next, we adjusted the parameters to control the effect of each term.

From several experimental trials, the learning rate of the World model, the Actor model, and the Critic model were set to $4\times 10^{-4}$, $4\times 10^{-4}$ and $2\times 10^{-4}$, respectively.


In the training phase of this experiment,
we used trajectories that move a random value within a range from $-25^{\circ}$ to $40^{\circ}$.
The training was conducted until the total reward during the evaluation was mostly at the maximum reward and the losses of Tracker converged.
In this experiment, the training took approximately $40$ minutes to complete.

\subsection{Result of tracking control}

\begin{table}[tb]
    \caption{Tracker and PID controller's results of tracking $10$ same target trajectories including linear and nonlinear}
    \label{tab:mse}
    \begin{center}
    \scalebox{1}{
    \begin{tabular}{c|c|c}\hline
     & Tracker & PID controller \\\hline\hline
    MSE & $86 \pm 36$ & $461 \pm 433$ \\
    \hline
    \end{tabular}
    }
    \end{center}
\end{table}

We used ten trajectories as evaluation trajectories: four trajectories moving by random values within different ranges every step, two trajectories at different constant velocities, two trajectories moving by different values per $20$ steps, two sin curves at different speeds.

The Mean Squared Error (MSE) of Tracker and PID controller are shown in \tabref{tab:mse}. MSE was calculated by the following equation:
\begin{equation}
\begin{split}
    \frac{1}{T} \sum_{t=1}^{T} ( \theta_t -  \hat{\theta}_{t} )^2
\end{split}
\end{equation}
$\theta_t$ and $\hat{\theta}_t$ are the observable angle and target angle at timestep $t$, respectively. $T$ is the length of the evaluation trajectory.
MSE of Tracker was considerably smaller than that of PID controller.

\begin{figure}[tb]
\begin{center}
    \includegraphics[width=1\columnwidth]{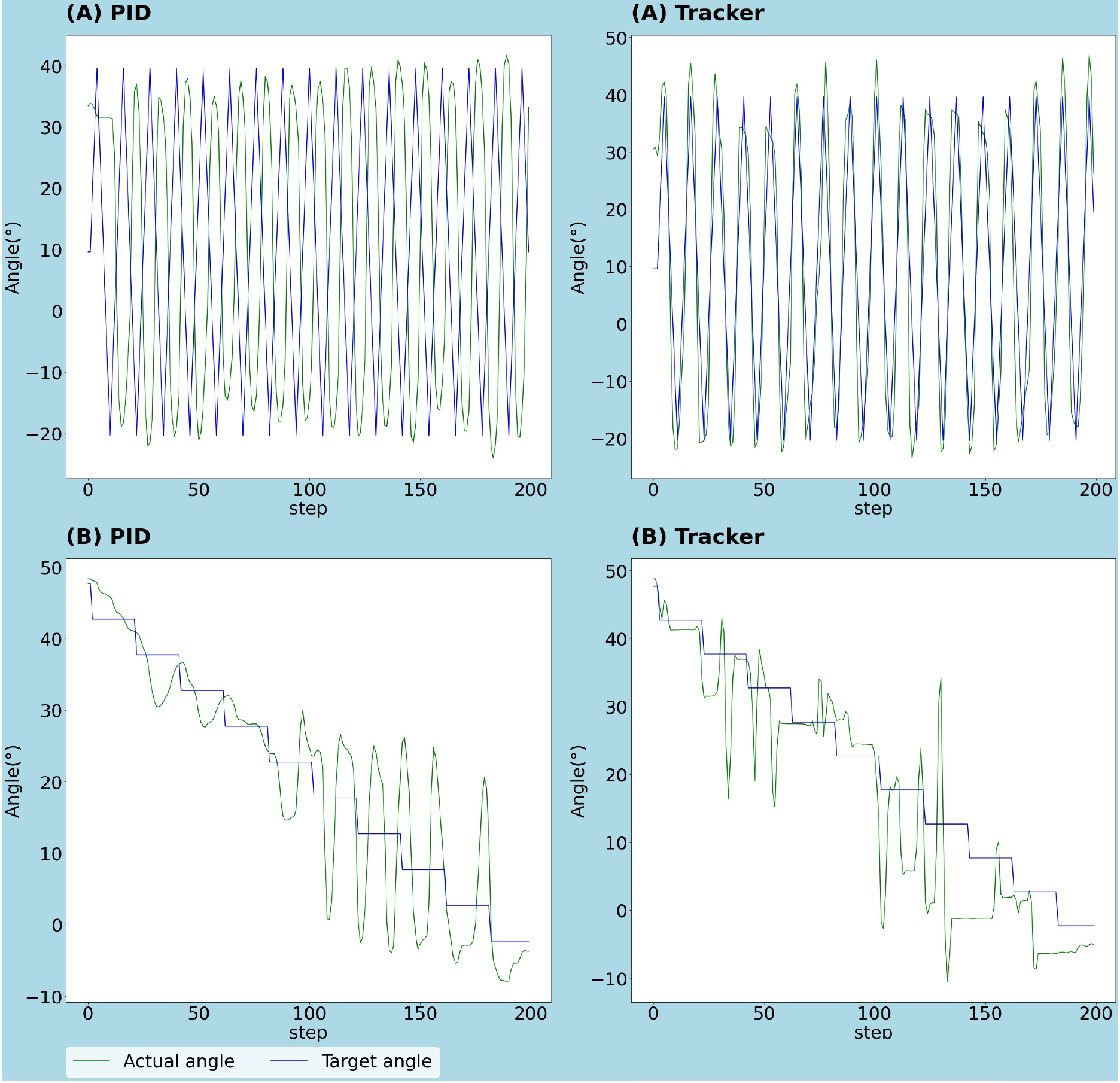}
    \caption{(A) The trajectory with the largest difference of MSE between our controller, Tracker, and PID in the evaluation trajectories (the one that moved by a constant velocity of $\pm 10^\circ$).
    (B) The trajectory with the smallest difference of MSE in the evaluation trajectories (the one that moved by $-5^\circ$ every $20$ steps).
    The left part is the result of PID controller and the right part is the result of our controller. }
    \label{fig:a}
\end{center}
\end{figure}

The trajectory with the largest difference of MSE was the one that moved by a constant velocity of $\pm 10^\circ$, where the MSE of our controller and PID controller were $102$ and $1440$ respectively (\figref{fig:a}-a). In contrast, the trajectory with the worst difference was the one that moved by $-5^\circ$ every $20$ steps, where the MSE of our controller and PID controller were $58$ and $48$ respectively (\figref{fig:a}-b). 
Because a trajectory moving in steps was not learned during training, Tracker could not control the joint. Therefore, the PID controller could achieve better results for this step trajectory. 
The standard deviation of the MSE was smaller for the proposed controller than for the PID controller (\tabref{tab:mse}). This indicates that our controller has the potential to track a wide range of trajectories than PID controller. In summary, this experimental result shows that Tracker can track more trajectories more accurately than the PID controller.

\section{Experiment I\hspace{-1.2pt}I}
To see if we could control a human finger with thin McKibben muscles, we used our controller to control the third joint of the human index finger.
In case that thin McKibben muscles are taken off and attached to human finger again, we should train the policy again because the condition and position of the muscles are slightly changed. It takes time and effort to train the policy from scratch. In this case, fine-tuning of the policy is generally effective for reducing the training cost. Thus, we also confirmed whether fine-tuning is possible using the trained model.

\subsection{Settings}
\subsubsection{Hardware settings}
This experiment was performed on the third joint ($n=1$) of the human index finger attached with thin McKibben muscles (\figref{fig:platform}-B). CobraGloves from Spice\footnote{https://mocap.jp/igs/} was used to acquire finger bending angles.
The data glove contained $16$ IMU sensors. The muscles were the same and placed in an antagonistic state as in \ref{ex1}.
In addition, to reduce the effects of gravity by changing in the wrist angle, the position and orientation of the wrist were fixed using acrylic plates.
We defined the joint angle at the fully extended finger as $0^\circ$ and the angle at finger flexion as a positive angle.

\subsubsection{Training settings}
 We used the same reward as that in \ref{ex1}. 
We preliminarily collected actual human finger motions as trajectories used in the training, and clipped these trajectories from $0^\circ$ to $40^\circ$ which is almost the range of joint motion realized by the used actuators. The trajectory dataset was collected such that the dataset included trajectories at four different speeds. In this experiment, as in \ref{ex1}, 
the training was conducted until the total reward during the evaluation was mostly at the maximum reward and the losses of Tracker converged.
In this experiment, it took approximately $60$ minutes to complete the training. After taking off and attaching the glove again for several minutes, the model was fine-tuned again using the trained model. This training took approximately $15$ minutes for all people. Nobody felt painful or awful during the training.

\subsubsection{Participants}
Individuals whose hands fit the data glove were recruited as participants. The experiment was conducted on three participants (mean age: 25 years, sex: female). They were instructed to relax their hands as much as possible during this experiment.

\subsection{Experimental results}

\subsubsection{Result of tracking control}

To investigate the effectiveness of Tracker, the trained policy was evaluated with four evaluation trajectories: step trajectory, slow trajectory, a little faster trajectory, fast trajectory. In the step trajectory, the joint angle increased in steps. In the other trajectory, the joint angle change at different speeds.
These trajectories were not used in the training.

\begin{figure}[tb]
\begin{center}
    \includegraphics[width=1\columnwidth]{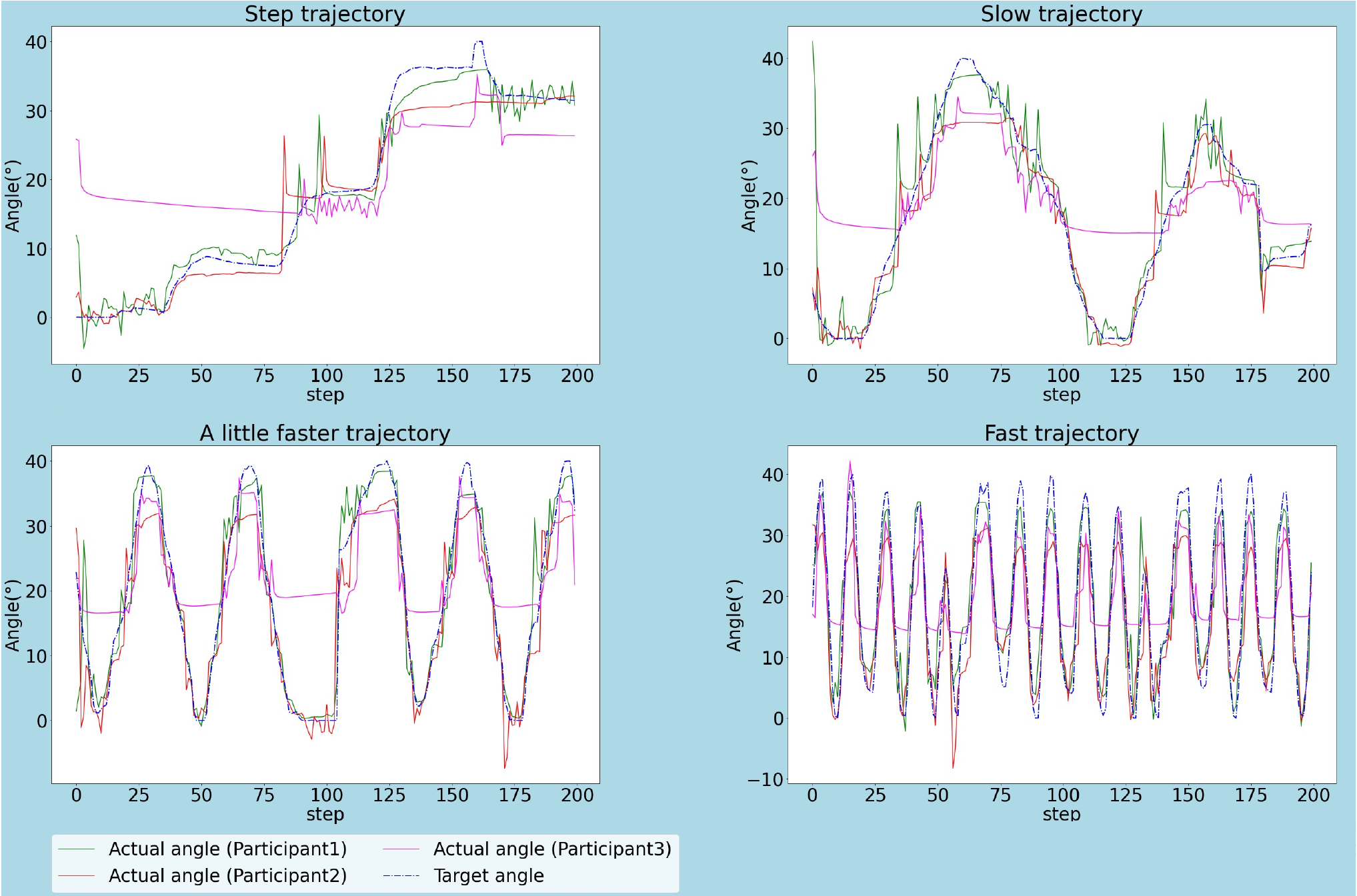}
    \caption{Four evaluation trajectories with different velocities and actual observations when moving the model obtained by the training. The green and red, pink lines represent the actual joint angles of participant 1 and 2, 3, respectively. The dotted blue line represents the target angles.}
    \label{fig:re}
\end{center}
\end{figure}

The results are shown in Table\ref{tab:finger} and Fig\ref{fig:re}. 
The MSEs of two participants were quite small, indicating that the tracking control of the human finger joint could be achieved with Tracker. The difference of MSE between participant 1 and participant 2 was also small. However, participant3's MSE was larger. Participant3 tended to have less finger movement than the other participants in the training phase. Also, participant 3 had better results because of the fine-tuning in the next experiment\ref{fine}. Therefore, it is possible that this could be improved by continuing the study for a longer time. The initialization of the policy is one of the main problems of reinforcement learning~\cite{andrychowicz2020matters}. The search of the good policy initialization is the future work.

MSE for the fast trajectory was larger. As the finger moves faster, a larger force is required to change the joint angle. Therefore, the optimal action is difficult to be found for the faster trajectory. We need to customize the actuators depending on the motion speed.

\begin{table}[tb]
    \centering
    \caption{Our controller's results of three participants for the four trajectories at different speeds.}
    \begin{tabular}{c|c|c|c}\hline
         MSE & Participant 1 & Participant 2 & Participant 3 \\\hline\hline
         First training & $15\pm 5.4$ & $19\pm 8.7$ & $83\pm 11$\\
        \hline
        After attaching again & $13\pm 7.4$& $32\pm 8.4$& $229\pm 18$\\
        \hline
        Fine-turning & $19\pm 7.0$& $14\pm 4.9$& $67\pm17 $\\
        \hline
    \end{tabular}
    \label{tab:finger}
\end{table}

\subsubsection{Result of fine-tuning}\label{fine}

After the policy was trained for approximately $60$ minutes, the thin McKibben muscles were taken off and attached gain. The policy was then fine-tuned for approximately $15$ minutes.

MSE of the fine-tuned policy was as shown in \tabref{tab:mse}. The same trajectories were used as those in the previous evaluation.
MSEs of participant 2 and 3 after fine-tuning decreased compared to MSE after attaching again. MSE of participant 1 increased, but is almost the same accuracy.
The result showed that it took approximately $15$ minutes, which is less than the time required for the first training, to achieve almost the same accuracy by fine-tuning the policy pre-trained by the user’s finger after taking off and attaching thin McKibben muscles again as the accuracy before taking off.

In this experiment, the policy was fine-tuned after several minutes from attaching the thin McKibben muscles again. Practically, there are cases that the policy should be fine-tuned after several days or months. Fine-tuning after a longer time is the future work.

\subsection{Result of World Model learning}

\begin{figure}[tb]
\begin{center}
\includegraphics[width=1\columnwidth]{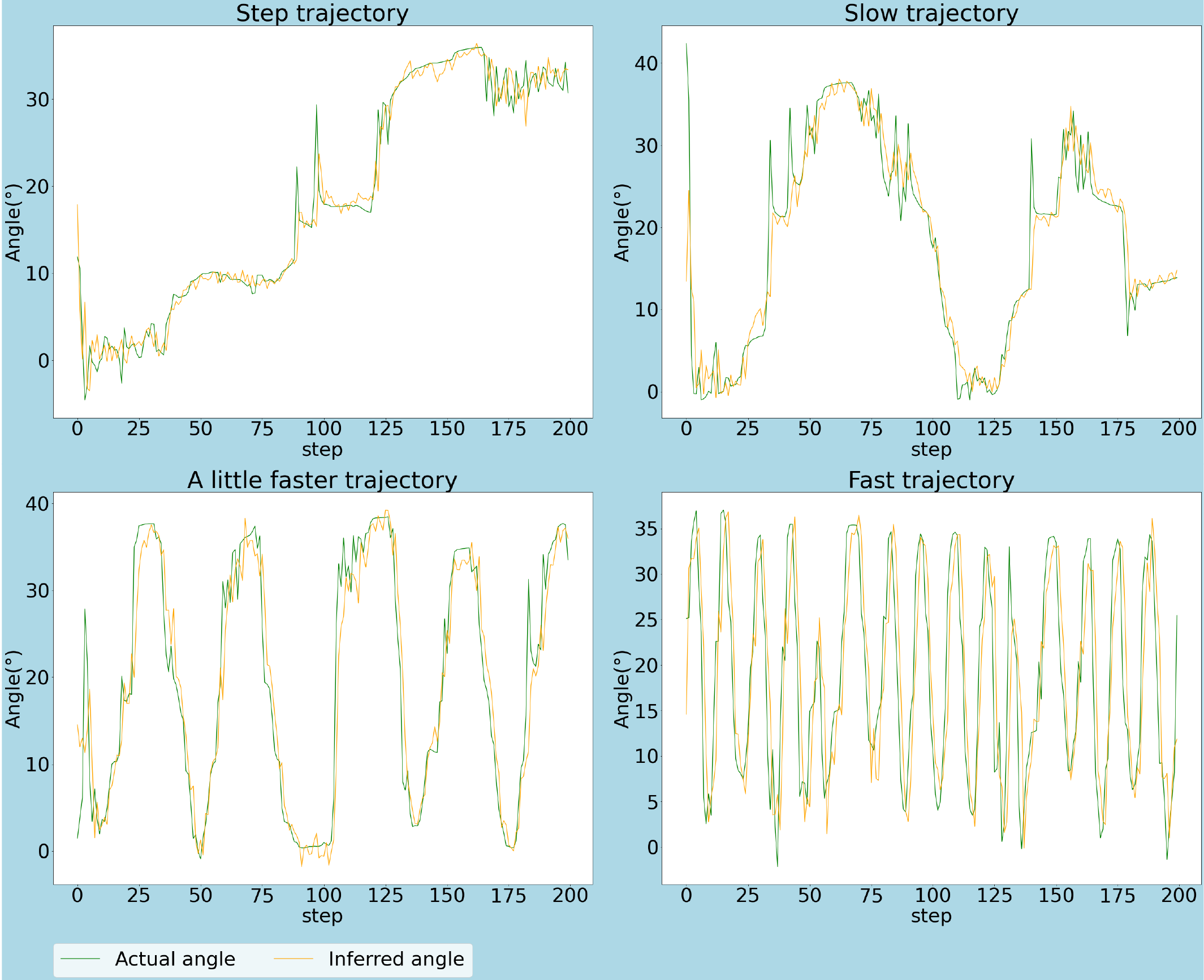}
    \caption{Joint angles inferred from the trained World Model 
    and actual joint angles.}
    \label{fig:world}
\end{center}
\end{figure}

We checked if World Model can be learned, which is the most important part of the policy training.
The joint angles were inferred from World Model using only Recurrent model and Transition model, Actor. The joint angles were inferred without actual joint angles because Representation model was not used.
The inferred and actual joint angles during the evaluation of Participant 1 were as shown in \figref{fig:world}.
When the actual joint angle decreases, the inferred joint angle also almost decreases. Similarly, when the actual joint angle increases, the inferred joint angle was also almost increases.
The result showed that World Model could roughly infer the joint angles of a human finger controlled by thin McKibben actuators. The successful learning of World Model should lead to the successful learning of the control policy.

\section{Discussion}
\subsection{Strength of a deep-neural-network based controller}
The experiment showed that Tracker, a deep-neural-network based controller, can achieve more precise control than the PID controller. Our deep-neural-network based controller has more parameters than the PID controller, so our controller can learn control commands more precisely. In addition, our controller can include not only errors but also future trajectories; thus, the forward-looking command can be decided. Our controller is more general to various trajectories than the PID controller because our controller can be learned using multiple pre-collected trajectories whereas the parameters of the PID controller are tuned using a single trajectory. 

\subsection{Considering state of human muscle toward the precise control}
In this study, we did not consider the state of human muscles, which should affect the control command. There is a possibility that the inclusion of the state of human muscles in the observation increases the control accuracy. A deep-neural-network based controller can input the muscle state, whereas the PID controller cannot. The deep-neural-network based controller has the advantage of integrating multimodal features, such as the joint angles, angle velocities, state of human muscles. The state of the human muscles can be obtained using electrical impedance tomography (EIT)~\cite{zhu2022muscle}.

\subsection{Time-efficient learning}
Our controller was trained using model-based reinforcement learning, which is a data-efficient learning method. However, it took an hour to learn the control policy. It will take more time to learn the policy for all fingers, which is a burden on users. To improve the usability, we need to accelerate the training. There is a possibility that the training speed is improved by fine-tuning pre-trained policy in a coarse simulator~\cite{truong2022rethinking} or using high-performance numerical computing library such as JAX~\cite{jax2018github, smith2022walk}. 

\section{Conclusion}
In this study, we proposed Tracker, which is an extension of DreamerV2, a model-based reinforcement learning method, for the tracking control problem. 
The experimental results showed that Tracker achieved more precise and general control for various trajectories than the PID controller for the two-link manipulator. In addition, Tracker achieved control of human fingers attached with thin McKibben muscles. 
It was also shown that World Model could be learned using thin McKibben muscles. This result suggested that World-Model-based reinforcement can be adopted not only for hard robots but also for soft robots.

Soft robots are suitable for human-robot interaction in terms of safety. This characteristic is synergistic with reinforcement learning on a real human and robot. The dynamics of the human hand vary from person to person, so we need to optimize an interactive system between human and robot for each individual. One of the solutions to solve this problem is to learn the dynamics from data on a real human and robot. This study is the key to solving the problem of the interaction between human and soft robot using reinforcement learning. 

The future direction of this work is to extend Tracker to all finger joints and interact with the objects using the soft hand exoskeleton controlled by Tracker, adopting Tracker to other soft materials.

\addtolength{\textheight}{-12cm}   




\section*{ACKNOWLEDGMENT}
This work was supported by JST CREST Grant Number JPMJCR17A3, Japan.

\bibliographystyle{unsrt}
\bibliography{bib}

\begin{thebibliography}{10}

\bibitem{pinto2020performance}
David Pinto-Fernandez, Diego Torricelli, Maria del Carmen Sanchez-Villamanan,
  Felix Aller, Katja Mombaur, Roberto Conti, Nicola Vitiello, Juan~C Moreno,
  and Jose~Luis Pons.
\newblock Performance evaluation of lower limb exoskeletons: a systematic
  review.
\newblock {\em IEEE Transactions on Neural Systems and Rehabilitation
  Engineering}, 28(7):1573--1583, 2020.

\bibitem{gull2020review}
Muhammad~Ahsan Gull, Shaoping Bai, and Thomas Bak.
\newblock A review on design of upper limb exoskeletons.
\newblock {\em Robotics}, 9(1):16, 2020.

\bibitem{du2021review}
Tiaan du~Plessis, Karim Djouani, and Christiaan Oosthuizen.
\newblock A review of active hand exoskeletons for rehabilitation and
  assistance.
\newblock {\em Robotics}, 10(1):40, 2021.

\bibitem{suzumori2015new}
Koichi Suzumori.
\newblock New pneumatic artificial muscle realizing giacometti robotics and
  soft robotics.
\newblock In {\em The 6th International Conference on Manufacturing, Machine
  Design and Tribology}, number~15, pages 4--5, 2015.

\bibitem{kurumaya2017design}
Shunichi Kurumaya, Hiroyuki Nabae, Gen Endo, and Koichi Suzumori.
\newblock Design of thin mckibben muscle and multifilament structure.
\newblock {\em Sensors and Actuators A: Physical}, 261:66--74, 2017.

\bibitem{wang2020sensor}
Biyuan Wang, Nobuhiro Takahashi, and Hideki Koike.
\newblock Sensor glove implemented with artificial muscle set for hand
  rehabilitation.
\newblock In {\em Proceedings of the Augmented Humans International
  Conference}, pages 1--4, 2020.

\bibitem{koizumi2020soft}
Shoichiro Koizumi, Te-Hsin Chang, Hiroyuki Nabae, Gen Endo, Koichi Suzumori,
  Motoki Mita, Kimio Saitoh, Kazutoshi Hatakeyama, Satoaki Chida, and Yoichi
  Shimada.
\newblock Soft robotic gloves with thin mckibben muscles for hand assist and
  rehabilitation.
\newblock In {\em 2020 IEEE/SICE International Symposium on System Integration
  (SII)}, pages 93--98. IEEE, 2020.

\bibitem{takahashi2020soft}
Nobuhiro Takahashi, Shinichi Furuya, and Hideki Koike.
\newblock Soft exoskeleton glove with human anatomical architecture: production
  of dexterous finger movements and skillful piano performance.
\newblock {\em IEEE Transactions on Haptics}, 13(4):679--690, 2020.

\bibitem{faudzi2016modeling}
Ahmad Athif~Mohd Faudzi, Noor Hanis Izzuddin~Mat Lazim, and Koichi Suzumori.
\newblock Modeling and force control of thin soft mckibben actuator.
\newblock {\em International Journal of Automation Technology}, 10(4):487--493,
  2016.

\bibitem{hafidz2022control}
Muhamad Hazwan~Abdul Hafidz, Nor Mohd~Haziq Norsahperi, Hong~Win Soon,
  Mohd~Najeb Jamaludin, and Ahmad Athif~Mohd Faudzi.
\newblock Control of thin mckibben muscles in an antagonistic pair
  configuration.
\newblock In {\em Computational Intelligence in Machine Learning: Select
  Proceedings of ICCIML 2021}, pages 155--163. Springer, 2022.

\bibitem{endo2020flexible}
Nobutsuna Endo, Yuta Kizaki, and Norihiro Kamamichi.
\newblock Flexible pneumatic bending actuator for a robotic tongue.
\newblock {\em Journal of Robotics and Mechatronics}, 32(5):894--902, 2020.

\bibitem{yamaguchi2020three}
Hiroki Yamaguchi, Yuki Funabora, Shinji Doki, and Kae Doki.
\newblock Three-dimensional deformation control system for fabric actuator.
\newblock In {\em 2020 IEEE/SICE International Symposium on System Integration
  (SII)}, pages 253--258. IEEE, 2020.

\bibitem{arulkumaran2017deep}
Kai Arulkumaran, Marc~Peter Deisenroth, Miles Brundage, and Anil~Anthony
  Bharath.
\newblock Deep reinforcement learning: A brief survey.
\newblock {\em IEEE Signal Processing Magazine}, 34(6):26--38, 2017.

\bibitem{fan2020humanoid}
Jianyin Fan, Jing Jin, and Qiang Wang.
\newblock Humanoid muscle-skeleton robot arm design and control based on
  reinforcement learning.
\newblock In {\em 2020 15th IEEE Conference on Industrial Electronics and
  Applications (ICIEA)}, pages 541--546. IEEE, 2020.

\bibitem{ha2018world}
David Ha and J{\"u}rgen Schmidhuber.
\newblock World models.
\newblock {\em arXiv preprint arXiv:1803.10122}, 2018.

\bibitem{hafner2020mastering}
Danijar Hafner, Timothy Lillicrap, Mohammad Norouzi, and Jimmy Ba.
\newblock Mastering atari with discrete world models.
\newblock {\em arXiv preprint arXiv:2010.02193}, 2020.

\bibitem{wu2022daydreamer}
Philipp Wu, Alejandro Escontrela, Danijar Hafner, Ken Goldberg, and Pieter
  Abbeel.
\newblock Daydreamer: World models for physical robot learning.
\newblock {\em arXiv preprint arXiv:2206.14176}, 2022.

\bibitem{buchler2016lightweight}
Dieter B{\"u}chler, Heiko Ott, and Jan Peters.
\newblock A lightweight robotic arm with pneumatic muscles for robot learning.
\newblock In {\em 2016 IEEE International Conference on Robotics and Automation
  (ICRA)}, pages 4086--4092. IEEE, 2016.

\bibitem{suzuki2018novel}
Ryuji Suzuki, Manabu Okui, Shingo Iikawa, Yasuyuki Yamada, and Taro Nakamura.
\newblock Novel feedforward controller for straight-fiber-type artificial
  muscle based on an experimental identification model.
\newblock In {\em 2018 IEEE International Conference on Soft Robotics
  (RoboSoft)}, pages 31--38. IEEE, 2018.

\bibitem{melkou2019high}
Lamia Melkou and Mustapha Hamerlain.
\newblock High order homogeneous sliding mode control for a robot arm with
  pneumatic artificial muscles.
\newblock In {\em IECON 2019-45th Annual Conference of the IEEE Industrial
  Electronics Society}, volume~1, pages 394--399. IEEE, 2019.

\bibitem{kawaharazuka2019dyamic}
Kento Kawaharazuka, Toru Ogawa, and Cota Nabeshima.
\newblock Dynamic task control method of a flexible manipulator using a deep
  recurrent neural network.
\newblock In {\em 2019 IEEE/RSJ International Conference on Intelligent Robots
  and Systems (IROS)}, pages 7695--7701, 2019.

\bibitem{zhao2020sim}
Wenshuai Zhao, Jorge~Pe{\~n}a Queralta, and Tomi Westerlund.
\newblock Sim-to-real transfer in deep reinforcement learning for robotics: a
  survey.
\newblock In {\em 2020 IEEE symposium series on computational intelligence
  (SSCI)}, pages 737--744. IEEE, 2020.

\bibitem{Hafner2020Dream}
Danijar Hafner, Timothy Lillicrap, Jimmy Ba, and Mohammad Norouzi.
\newblock Dream to control: Learning behaviors by latent imagination.
\newblock In {\em International Conference on Learning Representations}, 2020.

\bibitem{hafner2019learning}
Danijar Hafner, Timothy Lillicrap, Ian Fischer, Ruben Villegas, David Ha,
  Honglak Lee, and James Davidson.
\newblock Learning latent dynamics for planning from pixels.
\newblock In {\em International conference on machine learning}, pages
  2555--2565. PMLR, 2019.

\bibitem{cho2014learning}
Kyunghyun Cho, Bart Van~Merri{\"e}nboer, Caglar Gulcehre, Dzmitry Bahdanau,
  Fethi Bougares, Holger Schwenk, and Yoshua Bengio.
\newblock Learning phrase representations using rnn encoder-decoder for
  statistical machine translation.
\newblock {\em arXiv preprint arXiv:1406.1078}, 2014.

\bibitem{clevert2015fast}
Djork-Arn{\'e} Clevert, Thomas Unterthiner, and Sepp Hochreiter.
\newblock Fast and accurate deep network learning by exponential linear units
  (elus).
\newblock {\em arXiv preprint arXiv:1511.07289}, 2015.

\bibitem{kingma2014adam}
Diederik~P Kingma and Jimmy Ba.
\newblock Adam: A method for stochastic optimization.
\newblock {\em arXiv preprint arXiv:1412.6980}, 2014.

\bibitem{zhang2020asynchronous}
Yunzhi Zhang, Ignasi Clavera, Boren Tsai, and Pieter Abbeel.
\newblock Asynchronous methods for model-based reinforcement learning.
\newblock In {\em Conference on Robot Learning}, pages 1338--1347. PMLR, 2020.

\bibitem{andrychowicz2020matters}
Marcin Andrychowicz, Anton Raichuk, Piotr Sta{\'n}czyk, Manu Orsini, Sertan
  Girgin, Raphael Marinier, L{\'e}onard Hussenot, Matthieu Geist, Olivier
  Pietquin, Marcin Michalski, et~al.
\newblock What matters in on-policy reinforcement learning? a large-scale
  empirical study.
\newblock {\em arXiv preprint arXiv:2006.05990}, 2020.

\bibitem{zhu2022muscle}
Junyi Zhu, Yuxuan Lei, Aashini Shah, Gila Schein, Hamid Ghaednia, Joseph
  Schwab, Casper Harteveld, and Stefanie Mueller.
\newblock Musclerehab: Improving unsupervised physical rehabilitation by
  monitoring and visualizing muscle engagement.
\newblock In {\em Proceedings of the 35th Annual ACM Symposium on User
  Interface Software and Technology}, UIST '22, New York, NY, USA, 2022.
  Association for Computing Machinery.

\bibitem{truong2022rethinking}
Joanne Truong, Max Rudolph, Naoki Yokoyama, Sonia Chernova, Dhruv Batra, and
  Akshara Rai.
\newblock Rethinking sim2real: Lower fidelity simulation leads to higher
  sim2real transfer in navigation.
\newblock {\em arXiv preprint arXiv:2207.10821}, 2022.

\bibitem{jax2018github}
James Bradbury, Roy Frostig, Peter Hawkins, Matthew~James Johnson, Chris Leary,
  Dougal Maclaurin, George Necula, Adam Paszke, Jake Vander{P}las, Skye
  Wanderman-{M}ilne, and Qiao Zhang.
\newblock {JAX}: composable transformations of {P}ython+{N}um{P}y programs,
  2018.

\bibitem{smith2022walk}
Laura Smith, Ilya Kostrikov, and Sergey Levine.
\newblock A walk in the park: Learning to walk in 20 minutes with model-free
  reinforcement learning.
\newblock {\em arXiv preprint arXiv:2208.07860}, 2022.

\end{thebibliography}

\end{document}